\documentclass{article}
\usepackage{spconf,amsmath,graphicx}
\usepackage{amssymb}
\usepackage{subfig}
\usepackage{bm}

\title{JOINT CTC-ATTENTION BASED END-TO-END SPEECH RECOGNITION\\
USING MULTI-TASK LEARNING}
\name{Suyoun Kim$^1$ $^2$, Takaaki Hori$^1$, and Shinji Watanabe$^1$\thanks{The work is performed during Suyoun Kim is at MERL.}}
\address{$^1$ Mitsubishi Electric Research Laboratories (MERL)\\
$^2$ Carnegie Mellon University (CMU)}
\begin{document}
\ninept
%
\maketitle
\begin{abstract}
Recently, there has been an increasing interest in end-to-end speech recognition that directly transcribes speech to text without any predefined alignments. One approach is the attention-based encoder-decoder framework that learns a mapping between variable-length input and output sequences in one step using a purely data-driven method. The attention model has often been shown to improve the performance over another end-to-end approach, the Connectionist Temporal Classification (CTC), mainly because it explicitly uses the history of the target character without any conditional independence assumptions. However, we observed that the performance of the attention has shown poor results in noisy condition and is hard to learn in the initial training stage with long input sequences. This is because the attention model is too flexible to predict proper alignments in such cases due to the lack of left-to-right constraints as used in CTC. This paper presents a novel method for end-to-end speech recognition to improve robustness and achieve fast convergence by using a joint CTC-attention model within the multi-task learning framework, thereby mitigating the alignment issue. An experiment on the WSJ and CHiME-4 tasks demonstrates its advantages over both the CTC and attention-based encoder-decoder baselines, showing 5.4-14.6\% relative improvements in Character Error Rate (CER).

\end{abstract}
\section{Introduction}
\label{sec:intro}
End-to-end speech recognition is a recently proposed approach that directly transcribes speech to text without requiring predefined alignment between acoustic frames and characters \cite{graves2014towards, hannun2014deep, miao2015eesen, chorowski2014end, chorowski2015attention, chan2015listen, bahdanau2015end, lu2016training, chan2016online}. The traditional hybrid approach, Deep Neural Networks - Hidden Markov Models (DNN-HMM), factorizes the system into several components trained separately (i.e. acoustic model, context-dependent phone transducer, pronunciation model, and language model) based on conditional independence assumptions (including Markov assumptions) and approximations \cite{mohamed2012acoustic, hinton2012deep}. Unlike such hybrid approaches, the end-to-end model learns acoustic frames to character mappings in one step towards the final objective of interest, and attempts to rectify the suboptimal issues that arise from the disjoint training procedure.  

Recent work on end-to-end speech recognition can be categorized into two main approaches: Connectionist Temporal Classification (CTC) \cite{graves2006connectionist, graves2014towards, hannun2014deep, miao2015eesen} and attention-based encoder-decoder \cite{bahdanau2014neural, chorowski2014end, chorowski2015attention, chan2015listen}. Both methods address the problem of variable-length input and output sequences. The key idea of CTC is to use intermediate label representation allowing repetitions of labels and occurrences of blank labels to identify no output label. The CTC loss can be efficiently calculated by the forward-backward algorithm, but it still predicts targets for every frame, and assumes that the targets are conditionally independent of each other.

Another approach, the attention-based encoder-decoder directly learns a mapping from acoustic frame to character sequences. At each output time step, the model emits a character conditioned on the inputs and the history of the target character. Since the attention model does not use any conditional independence assumption, it has often shown to improve Character Error Rate (CER) than CTC when no external language model is used \cite{bahdanau2015end}. However, in real-environment speech recognition tasks, the model shows poor results because the alignment estimated in the attention mechanism is easily corrupted due to the noise. Another issue is that the model is hard to learn from scratch due to the misalignment on longer input sequences, and therefore a windowing technique is commonly used to limit the area explored by the attention mechanism \cite{bahdanau2015end}, but several parameters for windowing need to be determined manually depending on the training data.

To overcome the above misalignment issues, this paper proposes a novel end-to-end speech recognition method to improve performance and accelerate learning by using a joint CTC-attention model within the multi-task learning framework. The key to our approach is that we use a shared-encoder representation trained by both CTC and attention model objectives simultaneously.
We think that the weakness of the attention model is due to lack of left-to-right constraints as used in DNN-HMM and CTC, making it difficult to train the encoder network with proper alignments in the case of noisy data and/or long input sequences.
Our proposed method improves the performance by rectifying the alignment problem using the CTC loss function based on the forward-backward algorithm. Along with improving performance, our framework significantly speeds up learning with fast convergence. We evaluate our model on the WSJ and CHiME-4 tasks, and show that our system outperforms both the CTC and attention models in CER and learning speed. 

\vspace{-0.3cm}
\section{JOINT CTC-ATTENTION MECHANISM}
\vspace{-0.3cm}
\label{sec:model}
In this section, we review the CTC in Section \ref{sec:ctc} and the attention-based encoder-decoder  in Section \ref{sec:attention}, addressing the variable ($T$) length input frames, $\bm{x} = (x_1, \cdots, x_T)$, and $U$ length output characters, $\bm{y} = (y_1, \cdots, y_U)$, where $y_u \in \{1, \cdots, K\}$. 
$K$ is the number of distinct labels.
Then, our joint CTC-attention based end-to-end framework will be described in Section \ref{sec:mtl}. 

\subsection{Connectionist temporal classification (CTC)}
\label{sec:ctc}
The key idea of CTC \cite{graves2006connectionist} is to use intermediate label representation $\pi = (\pi _1, \cdots, \pi _T)$, allowing repetitions of labels and occurrences of a blank label ($-$), which represents the special emission without labels, i.e., $\pi_t \in \{1, \cdots, K\} \cup \{-\}$. 
CTC trains the model to maximize $P(\bm{y}|\bm{x})$, the probability distribution over all possible label sequences $\Phi(\bm{y'})$: 
    \begin{align}
    \label{e1}
	P(\bm{y}|\bm{x}) = & \sum_{\bm{\pi} \in \Phi(\bm{y'})}P(\bm{\pi}|\bm{x}), 
    \end{align}  
where $\bm{y'}$ is a modified label sequence of $\bm{y}$, which is made by inserting the blank symbols between each label and the beginning and the end for allowing blanks in the output (i.e., $\bm{y} = (c,a,t), \bm{y'} = (-,c,-,a,-,t,-)$). 

CTC is generally applied on top of Recurrent Neural Networks (RNNs). Each RNN output unit is interpreted as the probability of observing the corresponding label at particular time. The probability of label sequence $P(\bm{\pi}|\bm{x})$ is modeled as being conditionally independent by the product of the network outputs:  
    \begin{align}
    \label{e2}
	P(\bm{\pi}|\bm{x}) \approx & \prod_{t=1}^T P(\pi _t|\bm{x}) = \prod_{t=1}^T q_t(\pi_t)
    \end{align}  
where $q_t(\pi_t)$ denotes the softmax activation of $\pi_t$ label in RNN output layer $q$ at time $t$. 

The CTC loss to be minimized is defined as the negative log likelihood of the ground truth character sequence $\bm{y^*}$, i.e.
    \begin{align}
	\label{e4}
    \mathcal{L}_\text{CTC} \triangleq & -\ln P(\bm{y^*}|\bm{x}).
    \end{align}
The probability distribution $P(\bm{y}|\bm{x})$ can be computed efficiently using the forward-backward algorithm as
    \begin{align}
    \label{e3}
	P(\bm{y}|\bm{x}) = \sum_{u=1}^{|\bm{y'}|} \frac{\alpha_t(u) \beta_t(u)} {q_t(y'_u)},
    \end{align}
where $\alpha_t(u)$ is the forward variable, representing the total probability of all possible prefixes ($y'_{1:u}$) that end with the $u$-th label, and $\beta_t(u)$ is the backward variable of all possible suffixes ($y'_{u:U}$) that start with the $u$-th label.
The network can then be trained with standard backpropagation by taking the derivative of the loss function with respect to $q_t(k)$ for any $k$ label including the blank.

%
%

Since CTC does not explicitly model inter-label dependencies based on the conditional independence assumption in Eq.\eqref{e2}, there are limits to model character-level language information. Therefore, lexicon or language models are commonly incorporated, like the hybrid framework \cite{hannun2014deep, miao2015eesen}. 

\subsection{Attention-based encoder-decoder}
\label{sec:attention}

Unlike the CTC approach, the attention model directly predicts each target without requiring intermediate representation or any assumptions, improving CER as compared to CTC when no external language model is used \cite{bahdanau2015end}. The model emits each label distribution at $u$ conditioning on previous labels according to the following recursive equations:
	\begin{align}
	    \label{e5}
		P(\bm{y}|\bm{x})  = & \prod_u P(y_u | \bm{x}, y_{1:u-1})\\
	    \label{e6}
		\bm{h}       = & \text{ Encoder}(\bm{x}) \\
	    \label{e7}
		y_u  \sim & \text{ AttentionDecoder}(\bm{h}, y_{1:u-1}).
	\end{align}

The framework consists of two RNNs: \textit{Encoder} and \textit{AttentionDecoder}, so that it is able to learn two different lengths of sequences based on the cross-entropy criterion. \textit{Encoder} transforms $\bm{x}$, to high-level representation $\bm{h} = (h_1, \cdots, h_L)$ in Eq. (\ref{e6}), then \textit{AttentionDecoder} produces the probability distribution over characters, $y_u$, conditioned on $\bm{h}$ and all the characters seen previously $y_{1:u-1}$ in Eq. (\ref{e7}). 
$L$ is the number of the frames of \textit{Encoder} output, and $L < T$.
Here, a special start-of-sentence\text{(sos)}/end-of-sentence\text{(eos)} token is added to the target set, so that the decoder completes the generation of the hypothesis when \text{(eos)} is emitted.
The loss function of the attention model is computed from Eq.~\eqref{e5} as:
    \begin{align}
		\label{eloss}
		\mathcal{L}_\text{Attention} \triangleq - \ln P(\bm{y}^*|\bm{x}) = -\sum_u \ln P(y_u^*|\bm{x},y^*_{1:u-1})
    \end{align}
where $y^*_{1:u-1}$ is the ground truth of the previous characters.

The attention mechanism aids in the decoding procedure by integrating all the inputs $\bm{h}$ into $c_u$ based on their attention weight vectors $a_{u} \in \mathbb{R}_{+}^L$ over input $L$ identifying where to focus at output step $u$. 
The following equations represent how to compute $a_u$ and $c_u$:
	\begin{align}
	    \label{e8}
           e_{u,l} = & 
            \begin{cases}
               \text{content-based:}\\
               \qquad w^T\tanh(Ws_{u-1} + Vh_l + b)\\
               \text{location-based:}\\
		       \qquad f_u = F*a_{u-1} \\
               \qquad w^T\tanh(Ws_{u-1} + Vh_l + Uf_{u,l} + b)\\
            \end{cases}\\
	    \label{e9}
		a_{u,l} = & \frac{\exp(\gamma e_{u,l})}{\sum_l \exp(\gamma e_{u,l})}\\
	    \label{e10}
		c_u = & \sum_l a_{u,l} h_l
	\end{align}
where $w, W, V, F, U, b$ are trainable parameters, $s_{u-1}$ is the decoder state, $\gamma$ is the sharpening factor \cite{chorowski2015attention}, and * denotes convolution.

$a_{u}$ can be computed by the softmax of energy $e_{u,l}$ from two types of attention mechanisms: content-based and location-based in Eq. (\ref{e8}). Both depend on the decoder state, $s_{u-1}$, and the content of input, $h_l$. The location-based attention mechanism additionally uses convolutional feature vectors $f_{u,l}$ extracted from the previous attention $a_{u-1}$ by convolving with matrix $F$ along the time axis \cite{chorowski2015attention}.

With $c_u$, $s_{u-1}$, and $y_{u-1}$, the decoder generates next label $y_u$ and updates the state as:
    \begin{align}
        \label{e11}
        y_u \sim & \text{ Generate}(c_u, s_{u-1}) \\
	    \label{e12}
		s_u    = & \text{ Recurrency}(s_{u-1}, c_u, y_u),  
    \end{align}
where the Generate and Recurrency functions indicate a feed-forward network and a recurrent network, respectively.

In practice, the approach has two main issues. (1) The model is weak on noisy speech data. The attention model is easily affected by noises, and generates misalignments because the model does not have any constraint that guides the alignments be monotonic as in DNN-HMM and CTC. (2) Another issue is that it is hard to learn from scratch on larger input sequences via purely data-driven methods. To make training faster, the authors \cite{chorowski2015attention, bahdanau2015end} constrains the attention mechanism to only consider inputs within a narrow range. However, this modification may limit the model's capability to extract useful information from long character sequences. 

\subsection{Proposed model: Joint CTC-attention (MTL)}
\label{sec:mtl}
The idea of our model is to use a CTC objective function as an auxiliary task to train the attention model encoder within the multi-task learning (MTL) framework.
\begin{figure}[tb]
\begin{minipage}[b]{1.0\linewidth}
  \centering
  \centerline{\includegraphics[width=8.5cm]{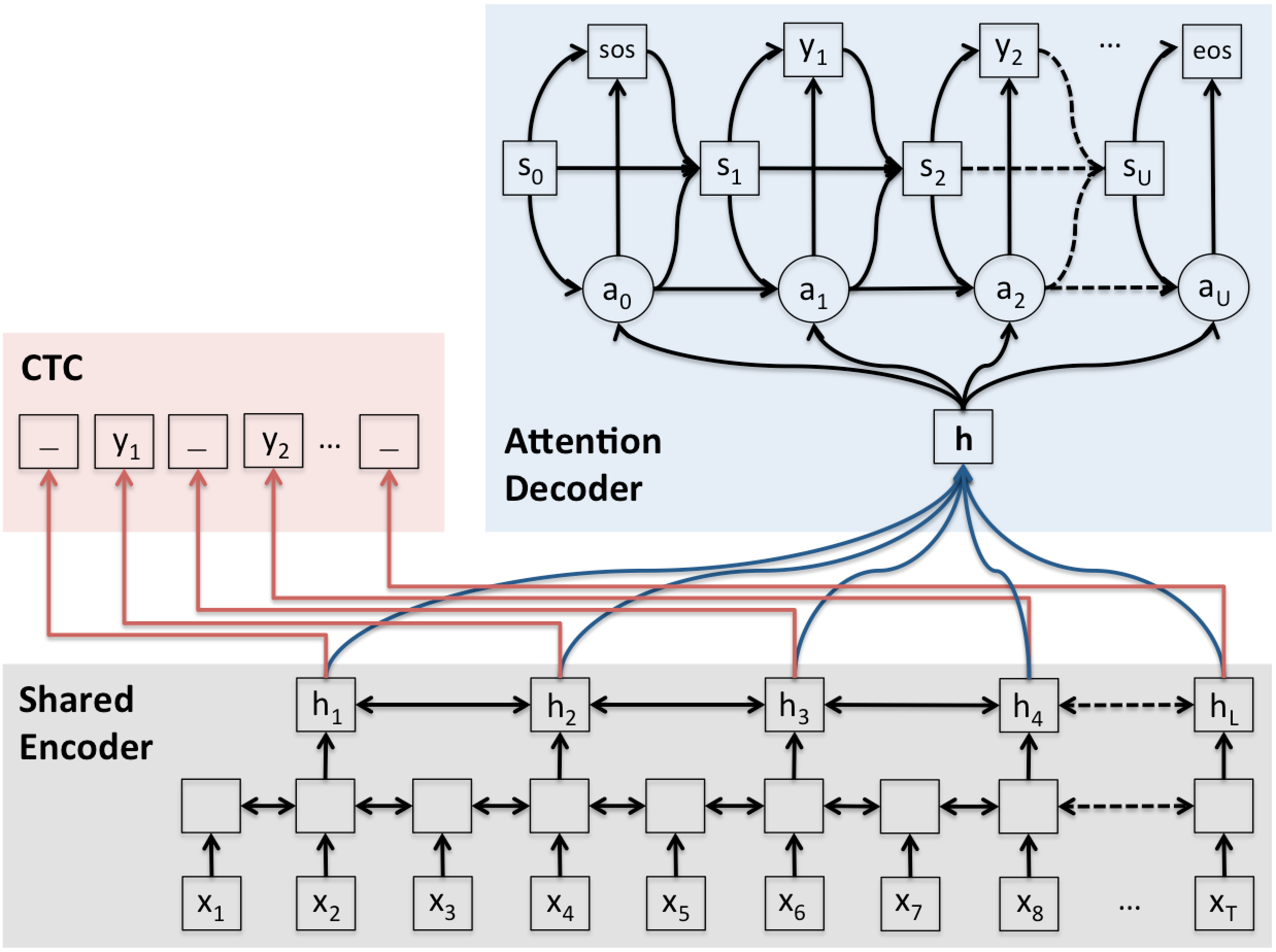}}
\end{minipage}
\caption{Our proposed Joint CTC-attention based end-to-end framework: the shared encoder is trained by both CTC and attention model objectives simultaneously. The shared encoder transforms our input sequence $\bm{x}$ into high level features $\bm{h}$, the location-based attention decoder generates the character sequence $\bm{y}$.}
\label{fig:arch}
\end{figure}
Figure \ref{fig:arch} illustrates the overall architecture of our framework, where the encoder network is shared with CTC and attention models. 
Unlike the attention model, the forward-backward algorithm of CTC can enforce monotonic alignment between speech and label sequences. We therefore expect that our framework is more robust in acquiring appropriate alignments in noisy conditions. Another advantage of using CTC as an auxiliary task is that the network is learned quickly. In our experiments, rather than solely depending on data-driven attention methods to estimate the desired alignments in long sequences, the forward-backward algorithm in CTC helps to speed up the process of estimating the desired alignment without the aid of rough estimates of the alignment which requires manual effort. 
The proposed objective is represented as follows by using both attention model in Eq.~\eqref{eloss} and CTC in Eq.~\eqref{e4}: 
	\begin{align}
	    \label{e12}
		\mathcal{L}_\text{MTL} &= \lambda \mathcal{L}_\text{CTC} + (1-\lambda) \mathcal{L}_\text{Attention},
	\end{align}
with a tunable parameter $\lambda: 0 \leq \lambda \leq 1$.

\section{EXPERIMENTS}
\label{sec:exp}

\subsection{Data}
We performed three sets of experiments: two on clean speech corpora, WSJ1 (81 hours) and WSJ0 (15 hours) \cite{wsj1, garofalo2007csr}, and one on a noisy speech corpus, CHiME-4 (18 hours) \cite{chime4}. The CHiME-4 corpus was recorded using a tablet device in everyday environments - a cafe, a street junction, public transport, and a pedestrian area. As input features, we used 40 mel-scale filterbank coefficients, with their first and second order temporal derivatives to obtain a total of 120 feature values per frame. Evaluation was done on (1) "eval92" for WSJ, and (2) "et05\_real\_isolated\_1ch\_track" for CHiME-4. Hyperparameter selection was performed on the (1) "dev93" for WSJ, and (2) "dt05\_multi\_isolated\_1ch\_track" for CHiME-4. None of our experiments used any language model or lexicon information. For the attention model, we used only 32 distinct labels: 26 characters, apostrophe, period, dash, space, noise, and sos/eos tokens. The CTC model uses the blank instead of sos/eos, and our MTL model uses both sos/eos and the blank.

\begin{table}
\caption{
Character Error Rate (CER) on clean corpora WSJ1 (80hours) and WSJ0 (15hours), and a noisy corpus CHiME-4 (18hours). None of our experiments used any language model or lexicon information. (Word Error Rate (WER) of our model MTL($\lambda=0.2$) was 18.2\% and WER of \cite{bahdanau2015end} was 18.6\% on WSJ1. Note that this is not an exact comparison because the hyper parameters were not completely same as \cite{bahdanau2015end}.)}
  \vspace{-0.5cm}
\label{tab:result}
\begin{center}
\begin{tabular}{r|c c}
  \hline
Model(train) 			& CER(valid)     & CER(eval) \\  \hline
  \hline
WSJ-train\_si284 (80hrs)			& dev93     & eval92 \\  \hline
CTC   		                    & 11.48	& 8.97 \\ 
Attention(content-based)      	& 13.68	& 11.08 \\
Attention(location-based)      	& 11.98	& 8.17 \\
MTL($\lambda=0.2$)   		    & {\bf 11.27} & {\bf 7.36} \\
MTL($\lambda=0.5$)   		    & 12.00 & 8.31 \\
MTL($\lambda=0.8$)   		    & 11.71 & 8.45 \\ \hline
  \hline
WSJ-train\_si84 (15hrs)			& dev93     & eval92 \\  \hline
CTC   		                    & 27.41 & 20.34 \\ 
Attention(content-based)      	& 28.02	& 20.06 \\
Attention(location-based)      	& 24.98	& 17.01 \\
MTL($\lambda=0.2$)   		    & {\bf 23.03}	& {\bf 14.53} \\
MTL($\lambda=0.5$)   		    & 26.28 & 16.24 \\
MTL($\lambda=0.8$)   		    & 32.21	& 21.30 \\ \hline
  \hline
CHiME-4-tr05\_multi (18hrs)		& dt05\_real & et05\_real \\  \hline
CTC   		                    & 37.56 & 48.79 \\ 
Attention(content-based)      	& 43.45	& 54.25 \\
Attention(location-based)      	& 35.01 & 47.58 \\
MTL($\lambda=0.2$)   		    & {\bf 32.08}	& {\bf 44.99} \\
MTL($\lambda=0.5$)   		    & 34.56	& 46.49 \\
MTL($\lambda=0.8$)   		    & 35.41	& 48.34 \\ \hline
\end{tabular}
\end{center}
\end{table}

\begin{figure}[!b]
\begin{minipage}[b]{1.0\linewidth}
  \centering
  \centerline{\includegraphics[width=8.5cm]{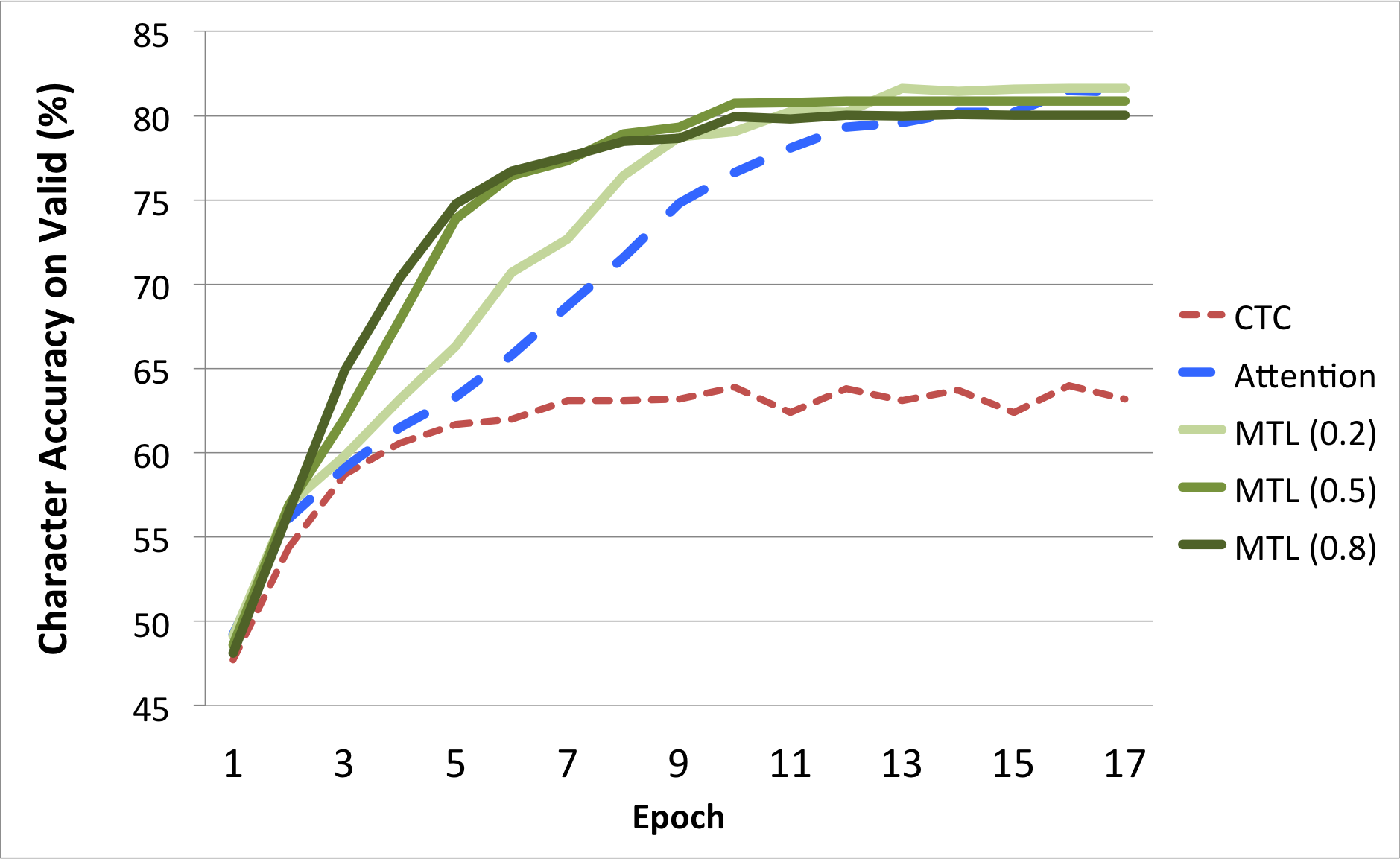}}
\end{minipage}
\caption{Comparison of learning curves: CTC, location-based attention model, and MTL with ($\lambda$ = 0.2, 0.5, 0.8). The character accuracy on the validation set of CHiME-4 is calculated by edit distance between hypothesis and reference. Note that the reference history were used in the attention and our MTL models.}
\label{fig:learningcurve}
\end{figure}

\begin{figure*}[!t]
\centering

\subfloat[Attention 1 epoch]{\includegraphics[width = 1.4in, height=1.4in]{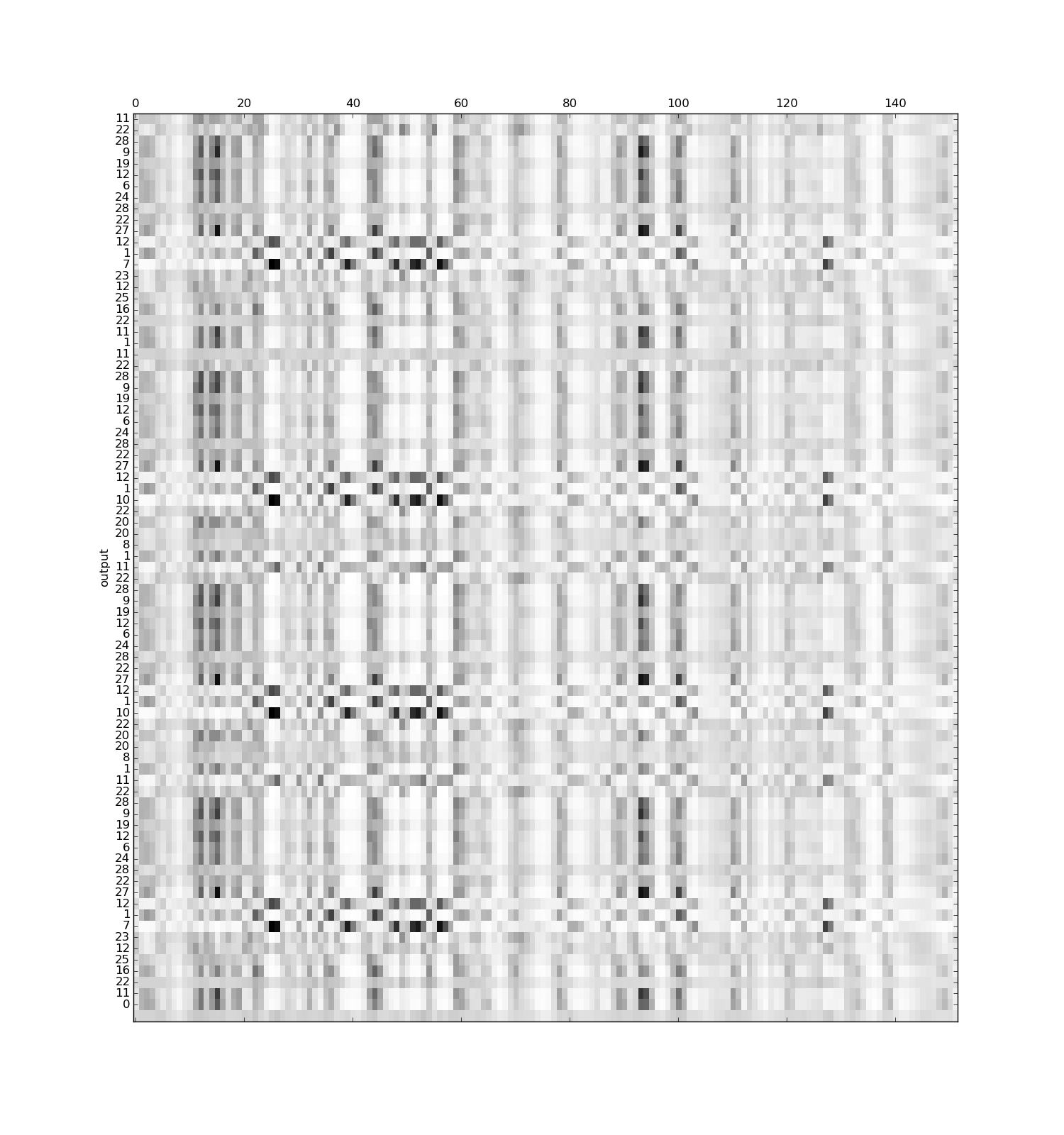}} 
\subfloat[Attention 3 epoch]{\includegraphics[width = 1.4in, height=1.4in]{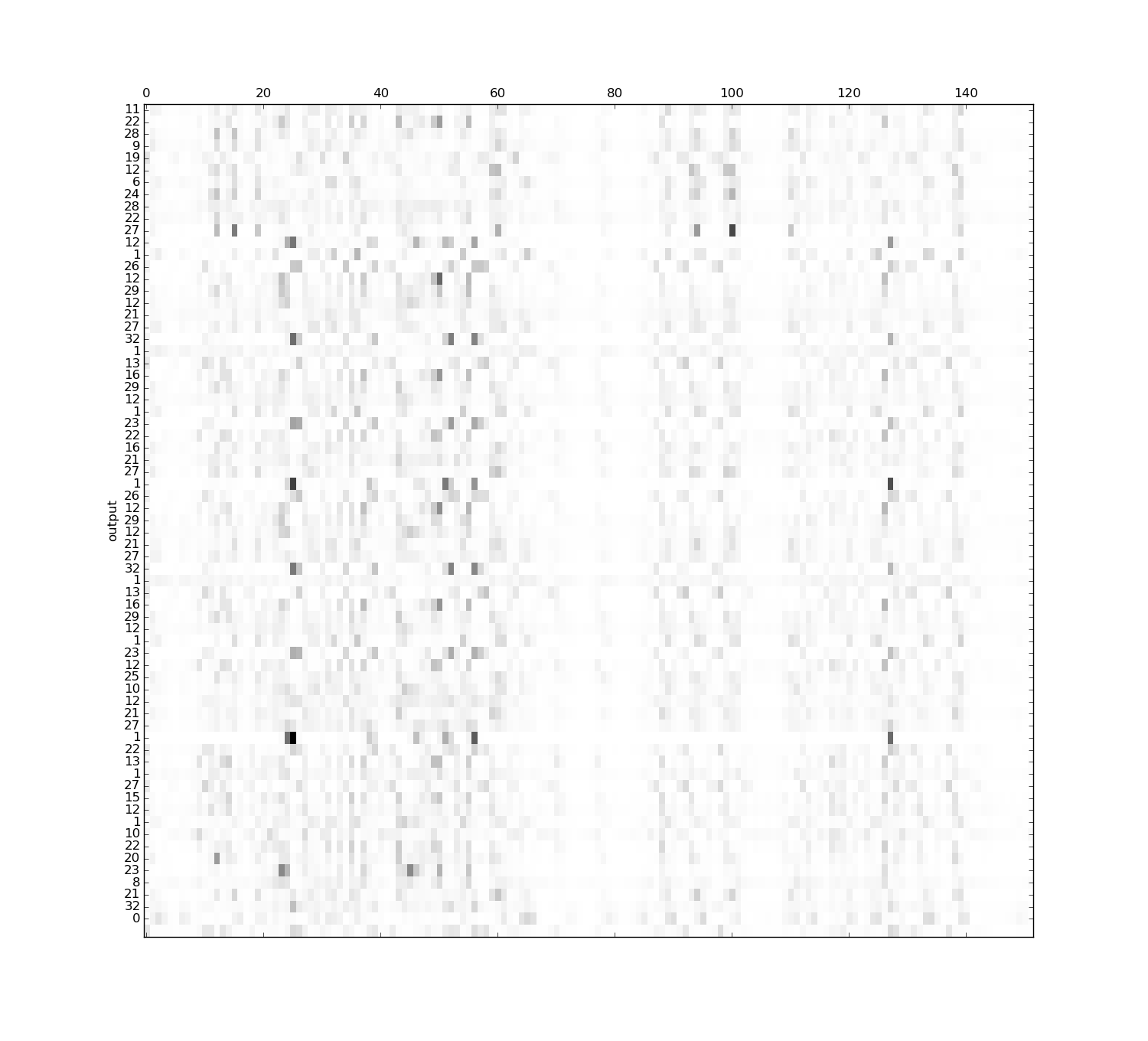}}
\subfloat[Attention 5 epoch]{\includegraphics[width = 1.4in, height=1.4in]{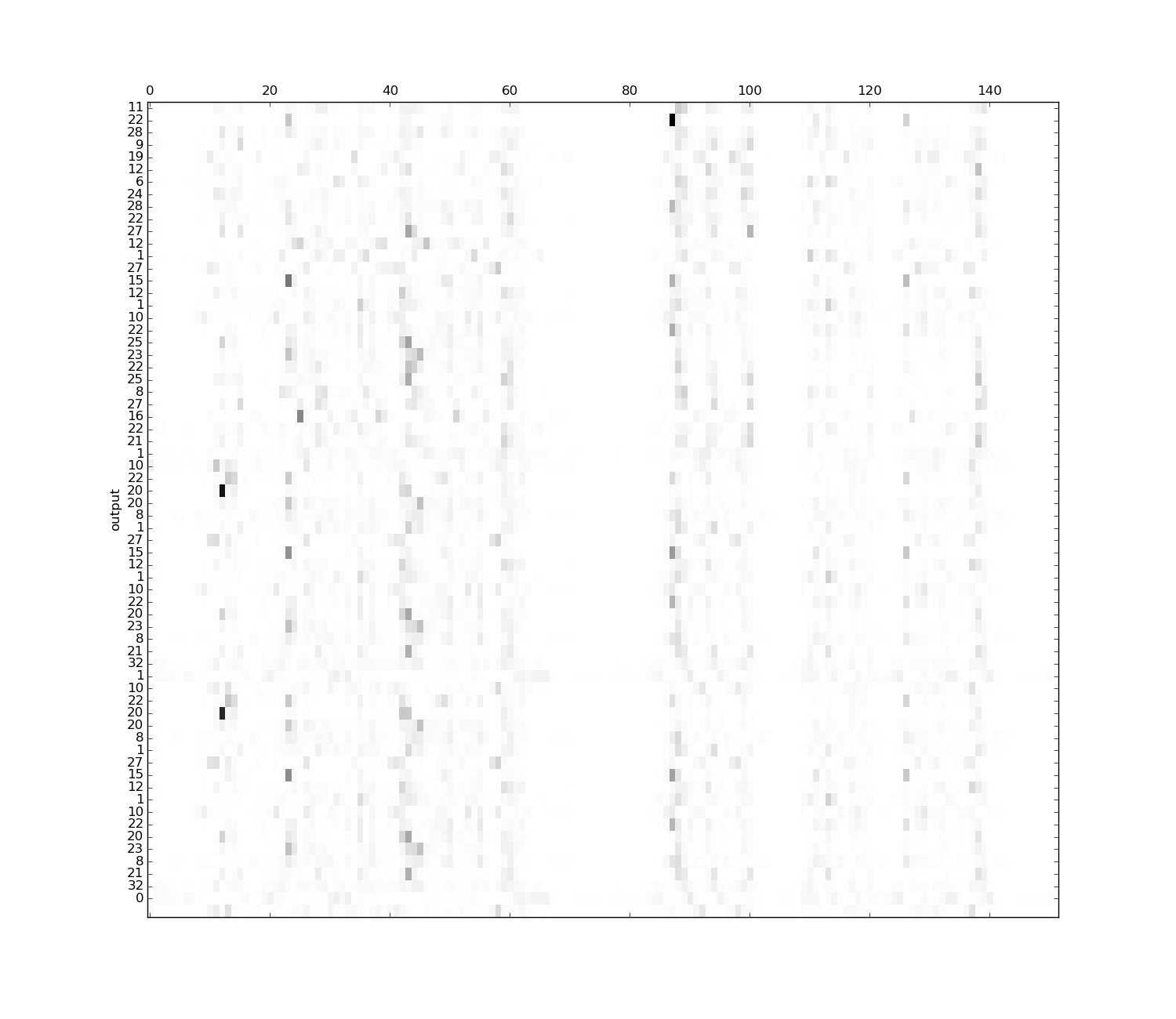}}
\subfloat[Attention 7 epoch]{\includegraphics[width = 1.4in, height=1.4in]{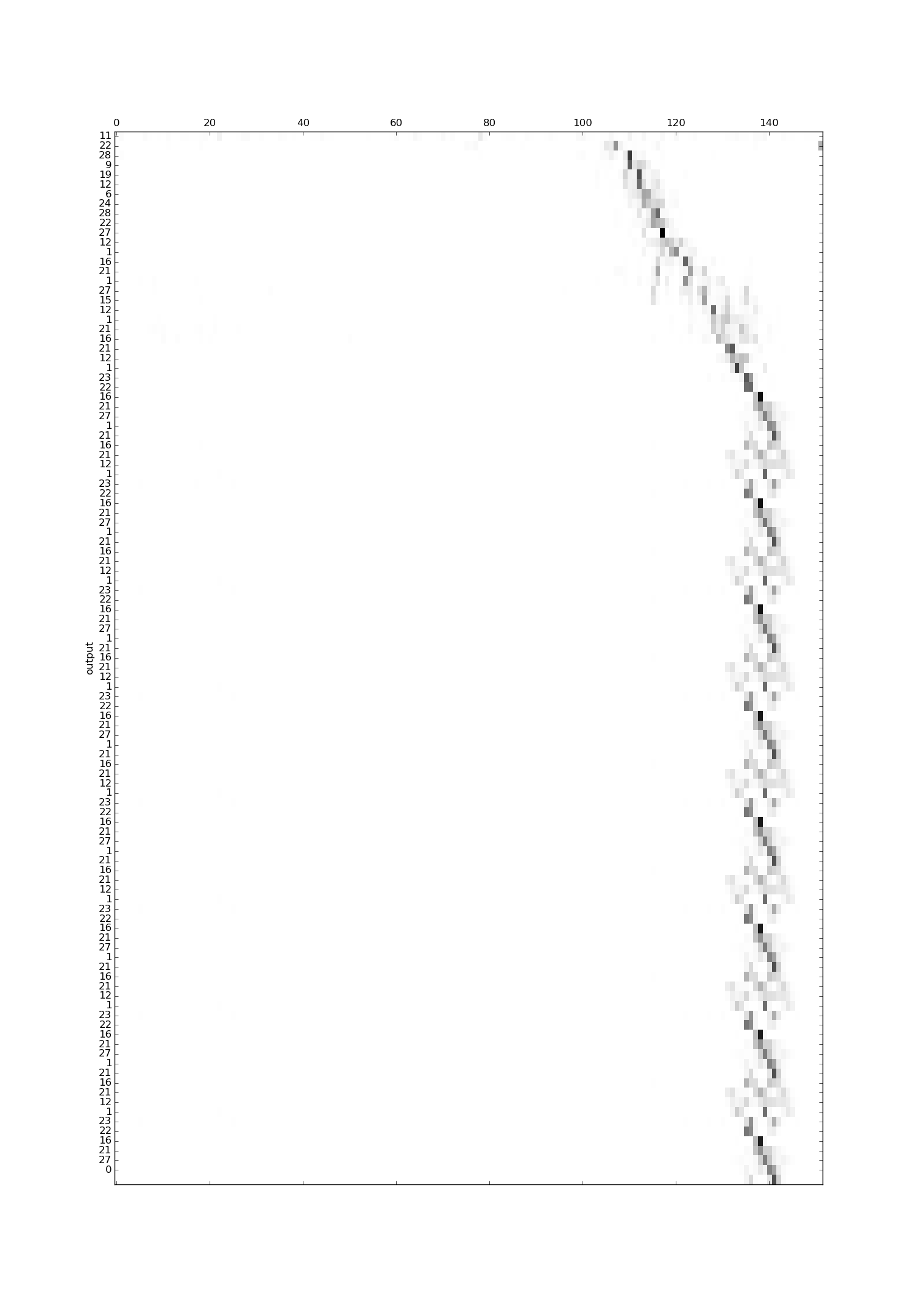}}
\subfloat[Attention 9 epoch]{\includegraphics[width = 1.4in, height=1.4in]{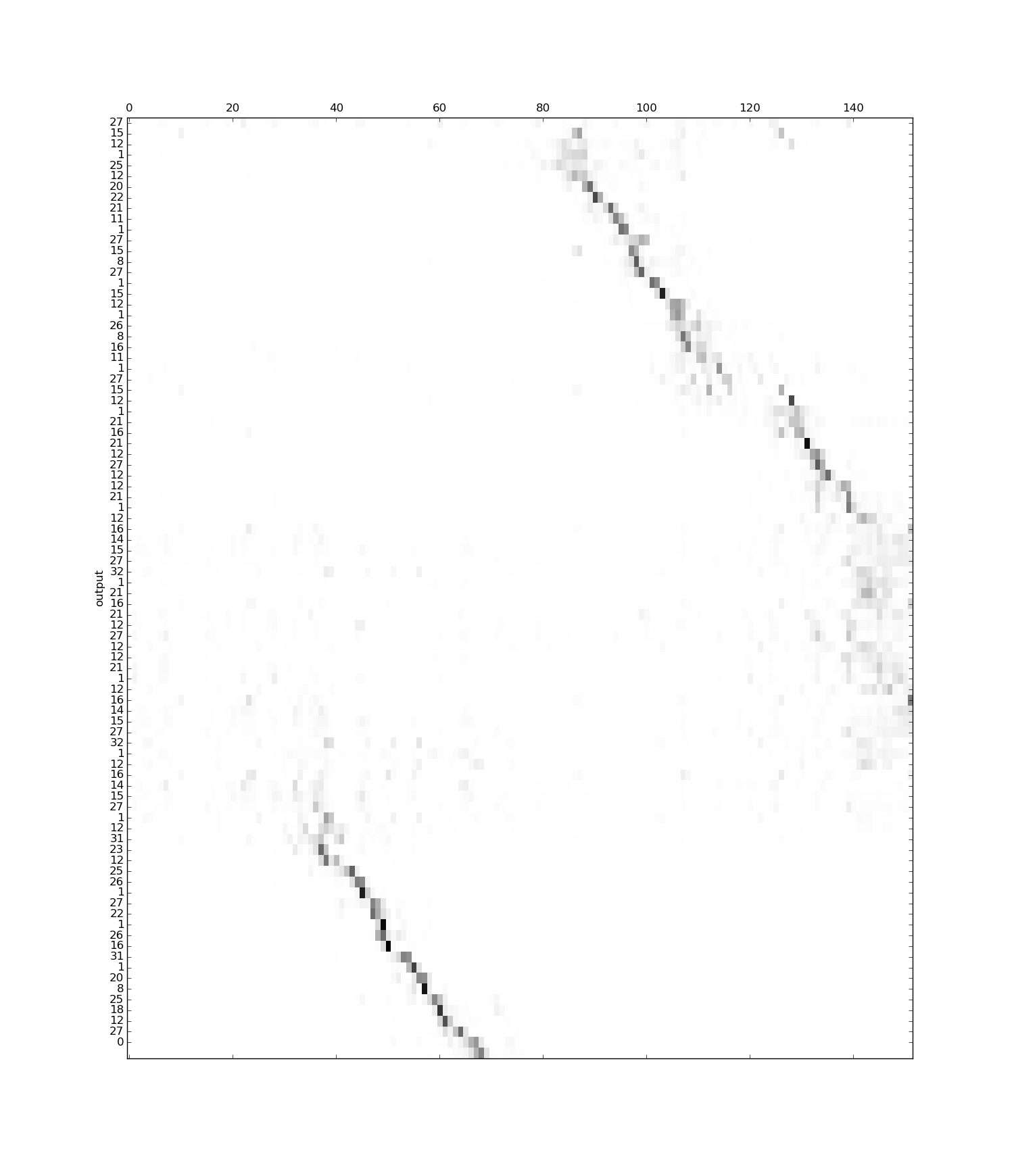}}
\\
\subfloat[MTL 1 epoch]{\includegraphics[width = 1.4in, height=1.4in]{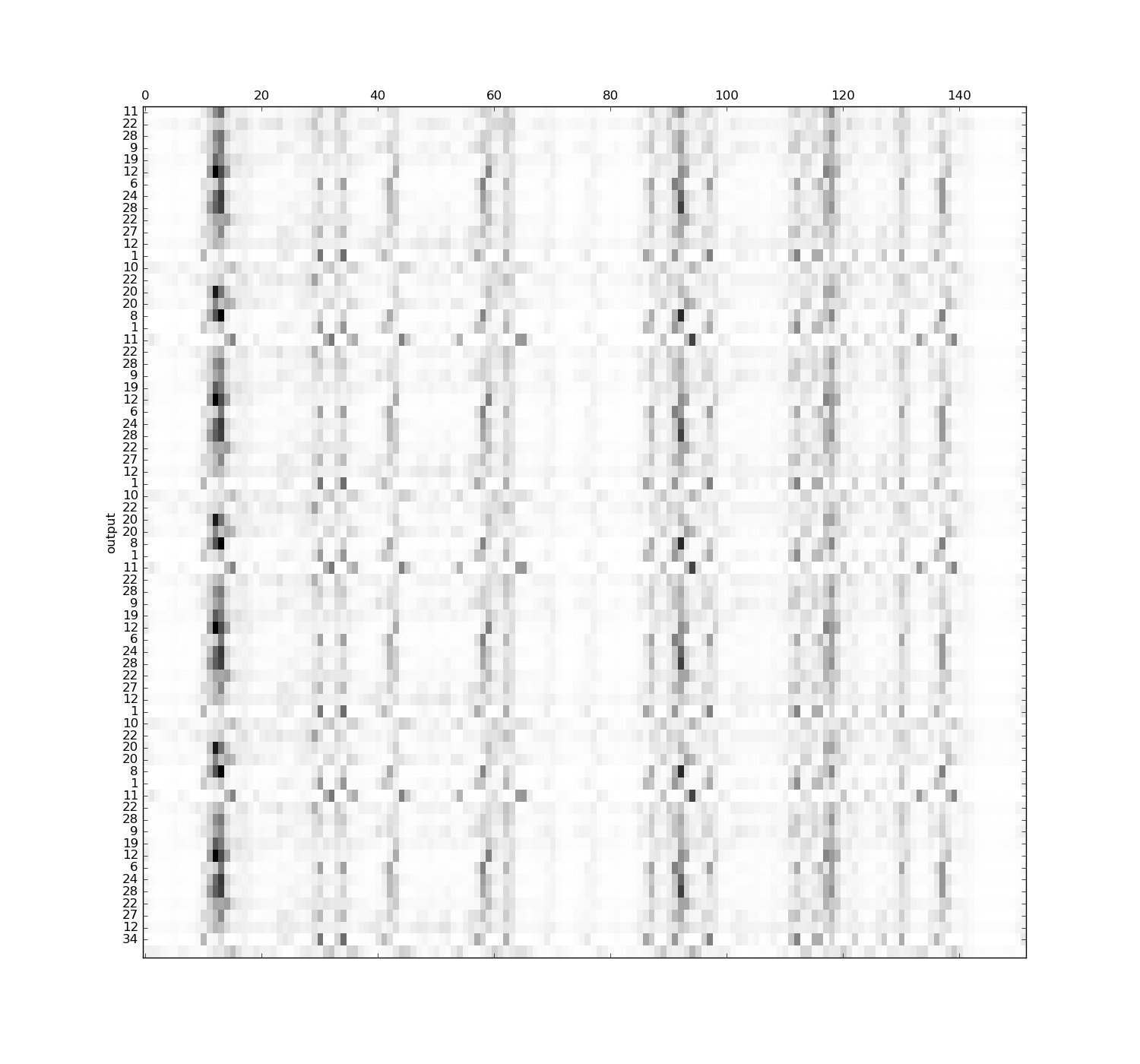}} 
\subfloat[MTL 3 epoch]{\includegraphics[width = 1.4in, height=1.4in]{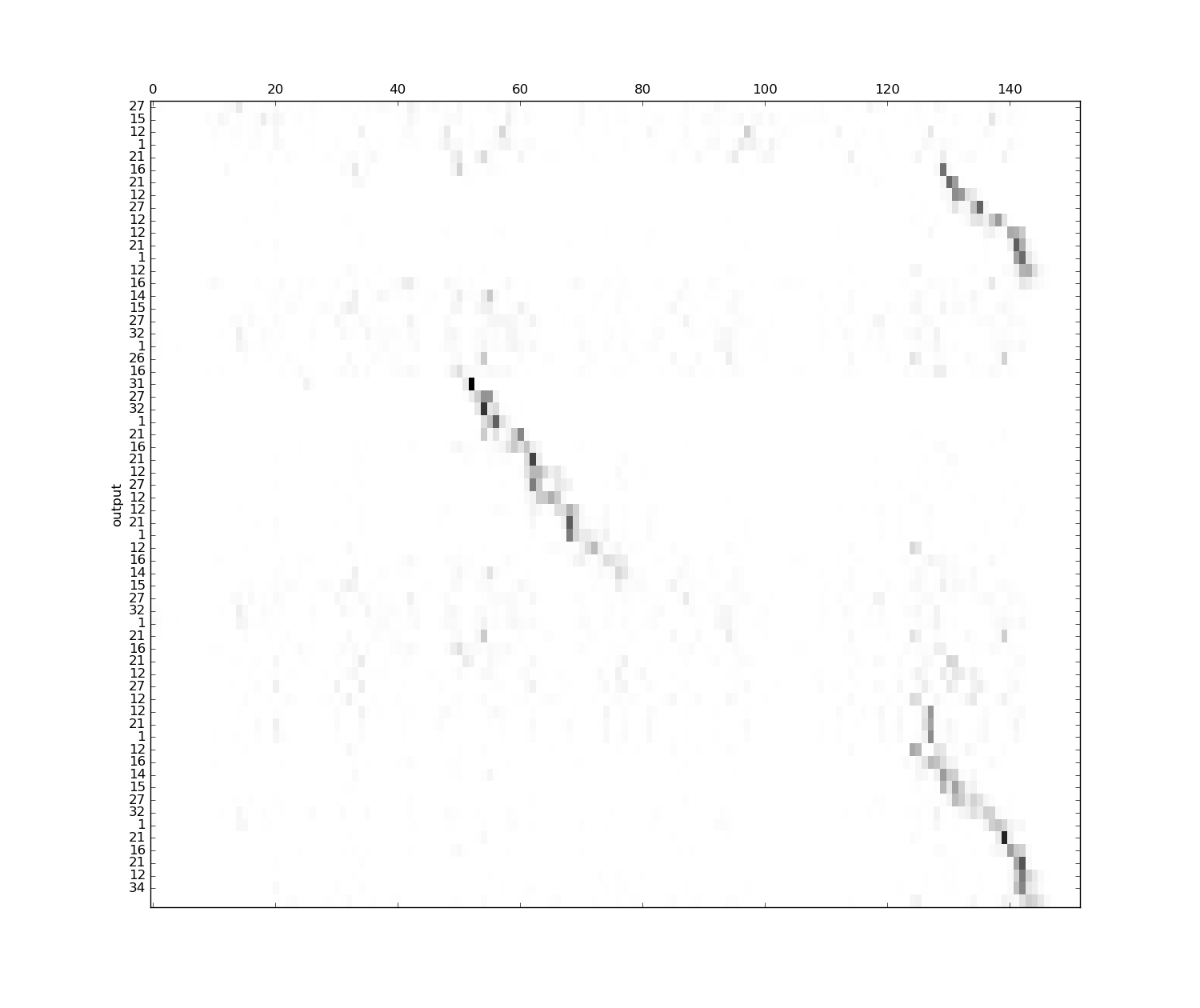}}
\subfloat[MTL 5 epoch]{\includegraphics[width = 1.4in, height=1.4in]{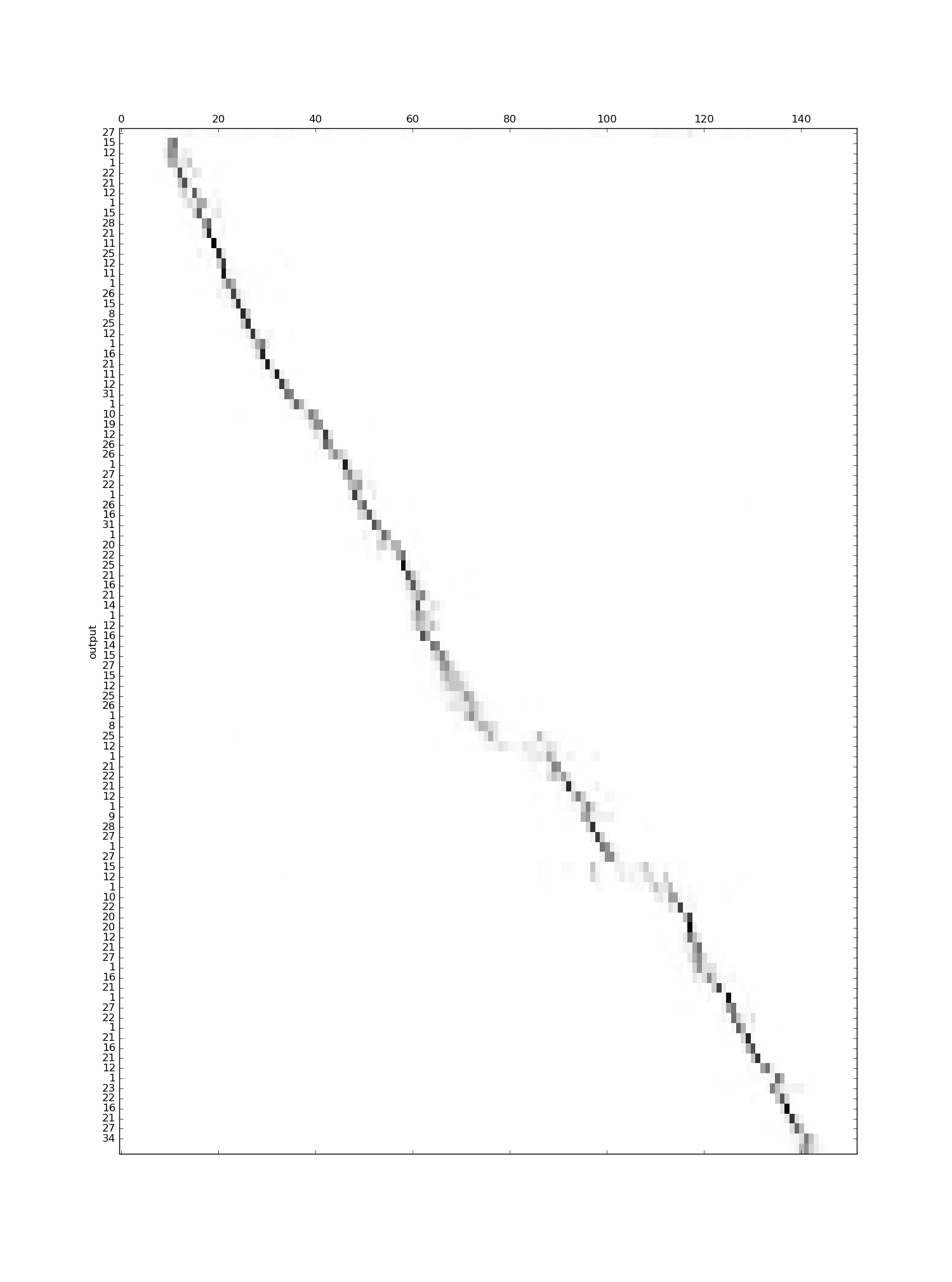}}
\subfloat[MTL 7 epoch]{\includegraphics[width = 1.4in, height=1.4in]{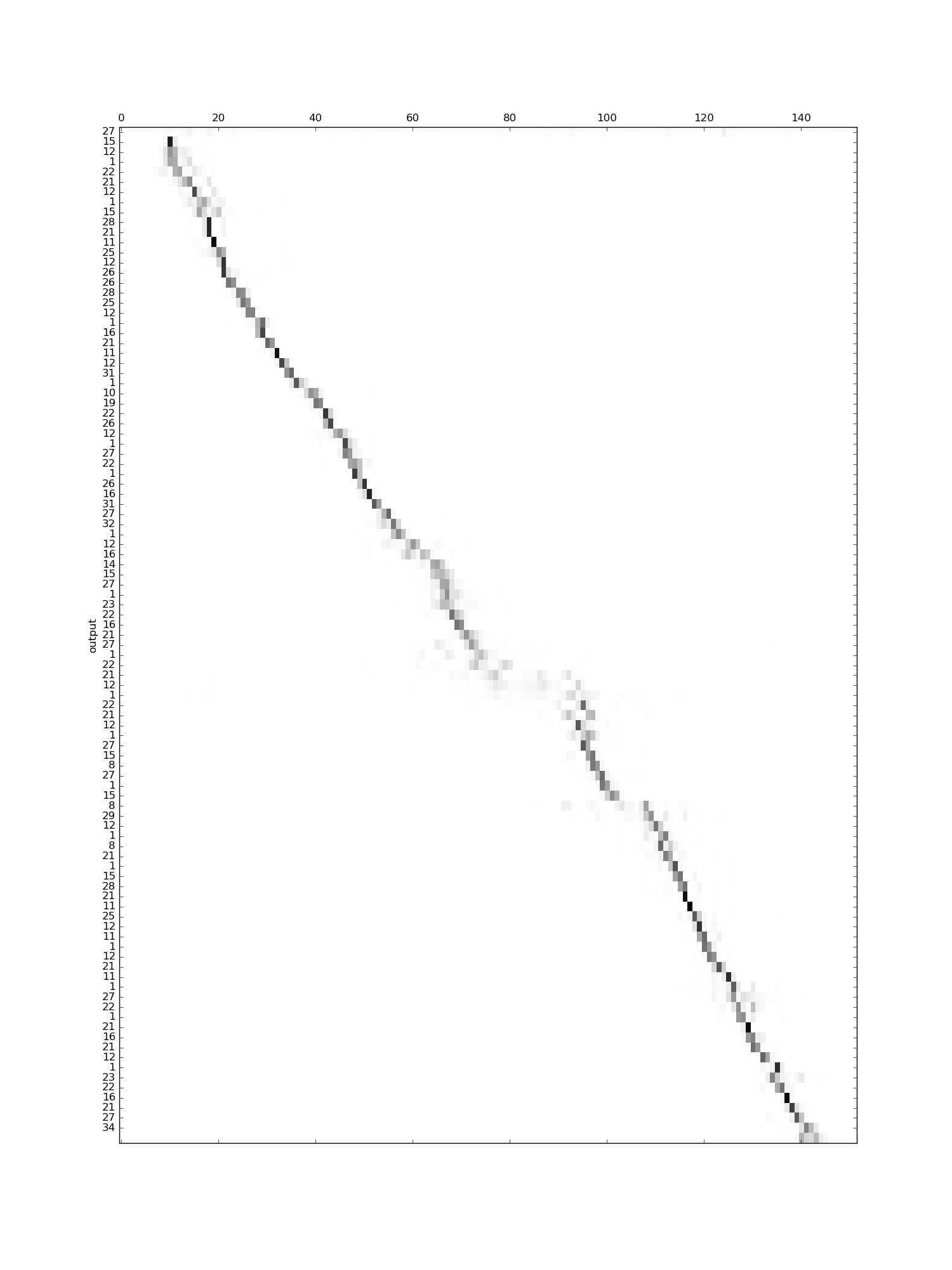}}
\subfloat[MTL 9 epoch]{\includegraphics[width = 1.4in, height=1.4in]{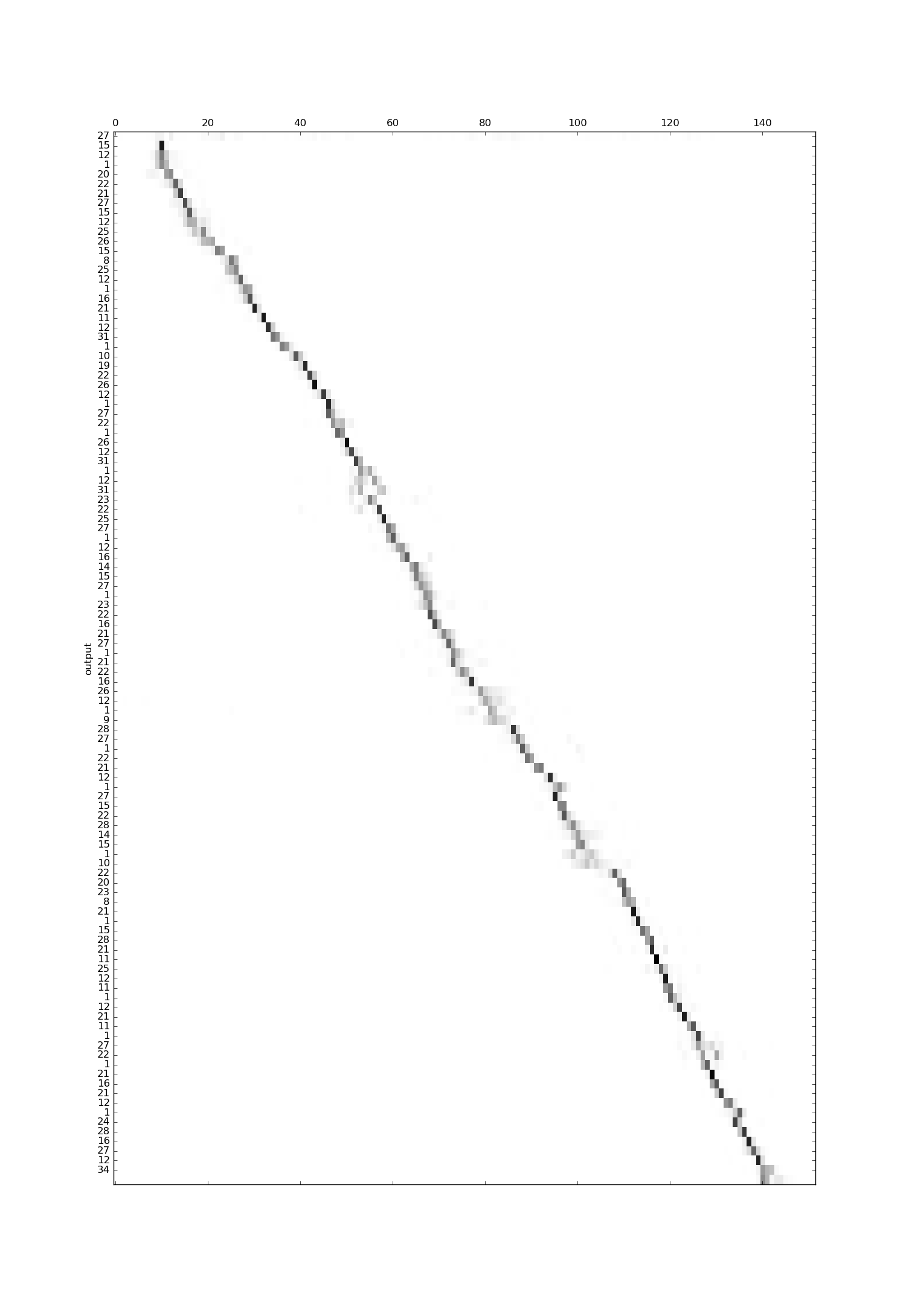}} 
\caption{Comparison of speed in learning alignments between characters (y-axis) and acoustic frames (x-axis) between the location-based attention model (1st row) and our model \texttt{MTL} (2nd row) over training epoch (1,3,5,7, and 9). All alignments are for one manually chosen utterance (F05\_442C020U\_CAF\_REAL - "THE ONE HUNDRED SHARE INDEX CLOSED SIX POINT EIGHT POINTS LOWER AT ONE THOUSAND SEVEN HUNDRED FIFTY NINE POINT NINE") in the noisy CHiME-4 evaluation set.} 
\label{fig:alignment}
\end{figure*}

\subsection{Training and Decoding}
The encoder was a 4-layer Bidirectional Long Short-Term Memory (BLSTM) \cite{hochreiter1997long, graves2013hybrid} with 320 cells in each layer and direction, and linear projection layer is followed by each BLSTM layer. The top two layers of the encoder read every second hidden state in the network below, reducing the utterance length by the factor of 4, $L=T/4$. The decoder was 1-layer LSTM with 320 cells. In case of the location-based attention model, 10 centered convolution filters of width 100 were used to extract the convolutional features. We used the sharpening factor $\gamma = 2$. The AdaDelta algorithm \cite{zeiler2012adadelta} with gradient clipping \cite{pascanu2012difficulty} was used for optimization. All the weights are initialized with the range [-0.1, 0.1] of uniform distribution. For our MTL, we tested three different task weights, $\lambda$: 0.2, 0.5, and 0.8.

For decoding of the attention and MTL models, we used a beam search algorithm similar to \cite{sutskever2014sequence} with the beam size 20 to reduce the computation cost. We adjusted the score by adding a length penalty, $\text{length(hyp)}*0.3$ for CHiME-4 and $\text{length(hyp)}*0.1$ for WSJ experiments. For decoding of CTC model, we took the sequence of most likely outputs. Note that we do not use any lexicon or language models. Our framework is implemented with the Chainer library \cite{tokui2015chainer,chainer}. 

\subsection{Results}
The results in Table \ref{tab:result} show that our proposed model MTL significantly outperformed both CTC and the attention model in CER on both the noisy CHiME-4 and clean WSJ tasks. Our model showed 6.0 - 8.4\% and 5.4 - 14.6\% relative improvements on validation and evaluation set, respectively. We observed that our joint CTC-attention achieved the best performance when we use the $\lambda = 0.2$ on both the noisy CHiME-4 and clean WSJ tasks.

One noticeable thing is that our framework significantly outperformed both the CTC and attention model even on clean corpora WSJ1 and WSJ0. It is possible that the CTC improved generalisation because of its training procedure that does not explicitly use character inter-dependencies. This point needs to be verified with additional experiments in future work.

Apart from the CER improvements, MTL can also be very helpful in accelerating the learning of the desired alignment. Figure \ref{fig:learningcurve} shows the learning curves of character accuracy on the validation sets of CHiME-4 over training epochs. Note that the accuracies of the attention and our MTL model were obtained with given gold standard history. As we use large $\lambda$ giving more weight to CTC loss, the network learns quickly and converges early. Figure \ref{fig:alignment} visualizes the attention alignments between characters and acoustic frames over training epoch. We observed that our MTL model learned the desired alignment in an early training stage, the 5th epoch, while the attention model could not learn the desired alignment even at the 9th epoch. 
This result indicates that the CTC loss guided the alignment to be monotonic in our MTL approach.

\section{CONCLUSIONS}
\label{sec:conclusion}

We have introduced a novel, general method for end-to-end speech recognition based on the multi-task learning approach using the CTC and the attention encoder-decoder. Our method improves performance by training a shared encoder using an auxiliary CTC objective function. Moreover, it significantly speeds up the process of learning the desired alignment without requiring manual restriction of the range of inputs, even in longer sequences. Our method has outperformed both CTC and an attention model on a speech recognition task in real-world noisy conditions as well as in clean conditions. This work can potentially be applied to any sequence-to-sequence learning task.

\vfill\pagebreak

\bibliographystyle{IEEEbib}
\bibliography{strings,refs}

\begin{thebibliography}{10}

\bibitem{graves2014towards}
Alex Graves and Navdeep Jaitly,
\newblock ``Towards end-to-end speech recognition with recurrent neural
  networks,''
\newblock in {\em Proceedings of the 31st International Conference on Machine
  Learning (ICML-14)}, 2014, pp. 1764--1772.

\bibitem{hannun2014deep}
Awni Hannun, Carl Case, Jared Casper, Bryan Catanzaro, Greg Diamos, Erich
  Elsen, Ryan Prenger, Sanjeev Satheesh, Shubho Sengupta, Adam Coates, et~al.,
\newblock ``Deep speech: Scaling up end-to-end speech recognition,''
\newblock {\em arXiv preprint arXiv:1412.5567}, 2014.

\bibitem{miao2015eesen}
Yajie Miao, Mohammad Gowayyed, and Florian Metze,
\newblock ``{EESEN}: End-to-end speech recognition using deep {RNN} models and
  {WFST}-based decoding,''
\newblock in {\em 2015 IEEE Workshop on Automatic Speech Recognition and
  Understanding (ASRU)}. IEEE, 2015, pp. 167--174.

\bibitem{chorowski2014end}
Jan Chorowski, Dzmitry Bahdanau, Kyunghyun Cho, and Yoshua Bengio,
\newblock ``End-to-end continuous speech recognition using attention-based
  recurrent {NN}: First results,''
\newblock {\em arXiv preprint arXiv:1412.1602}, 2014.

\bibitem{chorowski2015attention}
Jan~K Chorowski, Dzmitry Bahdanau, Dmitriy Serdyuk, Kyunghyun Cho, and Yoshua
  Bengio,
\newblock ``Attention-based models for speech recognition,''
\newblock in {\em Advances in Neural Information Processing Systems}, 2015, pp.
  577--585.

\bibitem{chan2015listen}
William Chan, Navdeep Jaitly, Quoc~V Le, and Oriol Vinyals,
\newblock ``Listen, attend and spell,''
\newblock {\em arXiv preprint arXiv:1508.01211}, 2015.

\bibitem{bahdanau2015end}
Dzmitry Bahdanau, Jan Chorowski, Dmitriy Serdyuk, Philemon Brakel, and Yoshua
  Bengio,
\newblock ``End-to-end attention-based large vocabulary speech recognition,''
\newblock {\em arXiv preprint arXiv:1508.04395}, 2015.

\bibitem{lu2016training}
Liang Lu, Xingxing Zhang, and Steve Renals,
\newblock ``On training the recurrent neural network encoder-decoder for large
  vocabulary end-to-end speech recognition,''
\newblock in {\em 2016 IEEE International Conference on Acoustics, Speech and
  Signal Processing (ICASSP)}. IEEE, 2016, pp. 5060--5064.

\bibitem{chan2016online}
William Chan and Ian Lane,
\newblock ``On online attention-based speech recognition and joint mandarin
  character-pinyin training,''
\newblock {\em Interspeech 2016}, pp. 3404--3408, 2016.

\bibitem{mohamed2012acoustic}
Abdel-rahman Mohamed, George~E Dahl, and Geoffrey Hinton,
\newblock ``Acoustic modeling using deep belief networks,''
\newblock {\em Audio, Speech, and Language Processing, IEEE Transactions on},
  vol. 20, no. 1, pp. 14--22, 2012.

\bibitem{hinton2012deep}
Geoffrey Hinton, Li~Deng, Dong Yu, George~E Dahl, Abdel-rahman Mohamed, Navdeep
  Jaitly, Andrew Senior, Vincent Vanhoucke, Patrick Nguyen, Tara~N Sainath,
  et~al.,
\newblock ``Deep neural networks for acoustic modeling in speech recognition:
  The shared views of four research groups,''
\newblock {\em Signal Processing Magazine, IEEE}, vol. 29, no. 6, pp. 82--97,
  2012.

\bibitem{graves2006connectionist}
Alex Graves, Santiago Fern{\'a}ndez, Faustino Gomez, and J{\"u}rgen
  Schmidhuber,
\newblock ``Connectionist temporal classification: labelling unsegmented
  sequence data with recurrent neural networks,''
\newblock in {\em Proceedings of the 23rd international conference on Machine
  learning}. ACM, 2006, pp. 369--376.

\bibitem{bahdanau2014neural}
Dzmitry Bahdanau, Kyunghyun Cho, and Yoshua Bengio,
\newblock ``Neural machine translation by jointly learning to align and
  translate,''
\newblock {\em arXiv preprint arXiv:1409.0473}, 2014.

\bibitem{wsj1}
Linguistic~Data Consortium,
\newblock ``{CSR}-{II} (wsj1) complete,''
\newblock {\em Linguistic Data Consortium, Philadelphia}, vol. LDC94S13A, 1994.

\bibitem{garofalo2007csr}
John Garofalo, David Graff, Doug Paul, and David Pallett,
\newblock ``{CSR}-{I} (wsj0) complete,''
\newblock {\em Linguistic Data Consortium, Philadelphia}, vol. LDC93S6A, 2007.

\bibitem{chime4}
Emmanuel Vincent, Shinji Watanabe, Aditya~Arie Nugraha, Jon Barker, and Ricard
  Marxer,
\newblock ``An analysis of environment, microphone and data simulation
  mismatches in robust speech recognition,''
\newblock in {\em Computer Speech and Language, to appear}.

\bibitem{hochreiter1997long}
Sepp Hochreiter and J{\"u}rgen Schmidhuber,
\newblock ``Long short-term memory,''
\newblock {\em Neural computation}, vol. 9, no. 8, pp. 1735--1780, 1997.

\bibitem{graves2013hybrid}
Alex Graves, Navdeep Jaitly, and Abdel-rahman Mohamed,
\newblock ``Hybrid speech recognition with deep bidirectional lstm,''
\newblock in {\em Automatic Speech Recognition and Understanding (ASRU), 2013
  IEEE Workshop on}. IEEE, 2013, pp. 273--278.

\bibitem{zeiler2012adadelta}
Matthew~D Zeiler,
\newblock ``Adadelta: an adaptive learning rate method,''
\newblock {\em arXiv preprint arXiv:1212.5701}, 2012.

\bibitem{pascanu2012difficulty}
Razvan Pascanu, Tomas Mikolov, and Yoshua Bengio,
\newblock ``On the difficulty of training recurrent neural networks,''
\newblock {\em arXiv preprint arXiv:1211.5063}, 2012.

\bibitem{sutskever2014sequence}
Ilya Sutskever, Oriol Vinyals, and Quoc~VV Le,
\newblock ``Sequence to sequence learning with neural networks,''
\newblock in {\em Advances in neural information processing systems}, 2014, pp.
  3104--3112.

\bibitem{tokui2015chainer}
Seiya Tokui, Kenta Oono, Shohei Hido, and Justin Clayton,
\newblock ``Chainer: a next-generation open source framework for deep
  learning,''
\newblock in {\em Proceedings of Workshop on Machine Learning Systems
  (LearningSys) in The Twenty-ninth Annual Conference on Neural Information
  Processing Systems (NIPS)}, 2015.

\bibitem{chainer}
Preferred Networks,
\newblock ``Chainer,''
\newblock in {\em "http://chainer.org/"}.

\end{thebibliography}

\end{document}